\documentclass[10pt,twocolumn,letterpaper]{article}

\usepackage[pagenumbers]{cvpr} 

\usepackage{cuted} 

\usepackage{currfile} 

\usepackage{caption} 

\usepackage{graphicx}
\usepackage{subcaption}
\usepackage{booktabs}
\usepackage{array}
\usepackage{multirow}
\usepackage{amsmath}
\usepackage{makecell}
\usepackage{threeparttable}
\usepackage{xcolor}
\usepackage{tabularx}
\graphicspath{{./}}
\usepackage[most]{tcolorbox}
\usepackage{xcolor}

\definecolor{PlannerBlue}{HTML}{416D9B}
\definecolor{SolverGreen}{HTML}{437C55}

\newtcolorbox{promptbox}[2]{
    enhanced,
    colback=#1!4!white,
    colframe=#1!75!white,
    colbacktitle=#1,
    coltitle=white,
    fonttitle=\small\bfseries,
    fontupper=\small,
    title={#2},
    attach boxed title to top left={
        xshift=3mm,
        yshift=-2mm
    },
    boxed title style={
        colback=#1,
        colframe=#1,
        boxrule=0pt,
        arc=1mm,
        left=2mm,
        right=2mm,
        top=0.5mm,
        bottom=0.5mm
    },
    boxrule=0.6pt,
    arc=2mm,
    left=2.5mm,
    right=2.5mm,
    top=3mm,
    bottom=2mm,
    before skip=8pt,
    after skip=8pt
}

\newcolumntype{Y}{>{\centering\arraybackslash}X} 

\definecolor{cvprblue}{rgb}{0.21,0.49,0.74}
\usepackage[pagebackref,breaklinks,colorlinks,allcolors=cvprblue]{hyperref}

\def\confName{CVPR}
\def\confYear{2027}

\title{Bridging the Modality Gap in KV-Cache Communication via Visually Conditioned Planning}

\author{%
Tiancheng Li$^{1}$ \quad
Qingyong Li$^{1}$\footnotemark[1] \quad
Minze Chen$^{2}$ \quad
Zhicheng Yan$^{1}$ \quad
Yang Liu$^{1}$ \quad
Wenju Sun$^{1}$\thanks{Correspondence to: Wenju Sun \textless sunwenju@bjtu.edu.cn\textgreater{} Qingyong Li \textless liqy@bjtu.edu.cn\textgreater{}} \\[0.5em]
$^{1}$School of Computer Science and Technology,
Beijing Jiaotong University \\
$^{2}$School of Mechatronics Engineering,
Beijing Institute of Technology \\[0.3em]
{\tt\small
    \{tianchengli,liqy,yanzhicheng,yliucit,sunwenju\}@bjtu.edu.cn
} \\
{\tt\small chenmz22@bit.edu.cn}
}

\begin{document}
\maketitle

\begin{abstract}

Inter-agent communication has become a key factor for the efficiency of growing multi-agent systems. KV-cache communication has recently emerged as a promising alternative, as it avoids the computational overhead of repeated prefill and enables layer-wise information propagation. However, we find that KV-cache communication severely degrades cross-modal performance in the VLM-to-LLM regime. We trace this failure to a persistent visual–language distribution gap in the transferred caches. This gap manifests as cross-modal attention decay, where the receiver under-attends to sender-provided visual states due to a mismatch between receiver queries and sender keys. Guided by this insight, we propose Conditioned-on Visual Language Cache Relay (Civil-Relay). Civil-Relay elicits a brief planning phase under visual context to encode additional information into the KV cache, which a lightweight projector then maps into the receiver’s decoding prefix. Across five multimodal benchmarks and two model scales, Civil-Relay consistently outperforms existing methods, improving average accuracy by 7.21 and 4.82 points over naive KV-cache projection at 4B and 8B, respectively.

\end{abstract}

\section{Introduction}


\begin{figure}[!t]
  \centering
  \includegraphics[width=0.9\linewidth]{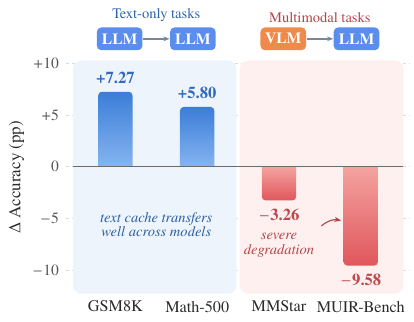}
  \caption{Accuracy differences between Naive Projection and Text-to-Text across language and multimodal tasks.
  Values denote Naive Projection minus Text-to-Text, in percentage points.}
  \label{fig:task_cmp}
\end{figure}


Large language model (LLM)-based multi-agent systems (MAS) have been increasingly adopted in practical applications~\cite{autogen2024,metagpt2024,chatdev2024,camel23iclr}. By assigning specialized roles to individual agents, such systems alleviate context pollution and facilitate capability specialization within each expert agent. However, as the systems continue to scale, both performance and overall efficiency increasingly hinge on the design of inter-agent communication.

The communication in existing multi-agent systems predominantly relies on natural language.
This language-based interface is inefficient, since the receiving agent requires an additional prefill pass to support downstream reasoning~\cite{promptcache24,cachegen24}, incurring extra computation and potentially introducing misinterpretation. To address these issues, some researchers transmit compact continuous states (e.g., expected embeddings, hidden activations, or compressed thoughts)~\cite{cipher2024,activation2025,interlat2026,condense2026,latcom2026}. Others transfer latent states when agents share a model backbone~\cite{latentmas2026,hylat2026}. The most information-preserving paradigm transmits information directly through the KV cache~\cite{c2c2025,dense2026}. 
Representative instantiations train lightweight projectors for corresponding layers across agents~\cite{c2c2025,dense2026,lcf2026,xkv2026,ucl26}. These projectors map the KV cache of a sending agent into a space compatible with the receiver's decoding process, enabling downstream reasoning to condition on the transferred states rather than re-encoded text. Despite its effectiveness, such lightweight projection relies on the implicit assumption that KV caches across different agents are drawn from transferable distributions.

The transferable assumption on KV cache may not hold in real-world settings. For highly specialized agents, they produce task-specific features and intricate attention patterns that could be difficult to transfer through simple transformations. To examine this issue, we consider a representative specialized MAS, the VLM-to-LLM regime, where a sending VLM handles image-domain inputs and a receiving LLM performs downstream reasoning. This cross-modal division preserves agent specialization while offering a controlled setting to isolate the effect of KV cache heterogeneity on inter-agent communication.

As shown in Figure~\ref{fig:task_cmp}, directly projecting the KV cache causes severe degradation in the VLM-to-LLM regime, whereas performance on text-only tasks remains largely preserved. To understand this phenomenon, we conduct a systematic analysis. Our visualization demonstrates that the KV caches of language tokens share a similar distribution across the two models, whereas visual and language tokens exhibit substantial distributional divergence. This visual–language gap persists even after lightweight projection, suggesting that the failure is not merely a projector misalignment but a modality-specific representational separation.

Further analysis reveals that this modality gap manifests in attention allocation. The receiver assigns disproportionately little attention to sender-relayed visual states relative to its own textual context. 
We term this phenomenon \textbf{cross-modal attention decay}. The attention decay stems from the mismatch between receiver queries and sender keys. Based on this analysis, we hypothesize that \textit{the key bottleneck of KV cache communication lies not in projector alignment but in whether the receiver allocates sufficient attention to sender-provided cross-modal states.} Improving attention allocation is therefore a promising direction for effective KV-cache transfer.

Inspired by this insight, we propose \textbf{C}ond\textbf{i}tioned-on \textbf{V}\textbf{i}sua\textbf{l} Language Cache Relay (Civil-Relay), a simple yet effective interface for cross-modal KV-cache communication. Civil-Relay mitigates cross-modal attention decay by relaying a visually conditioned language cache formed during a planning phase. Specifically, before transmitting the KV cache, the sender is elicited to perform a short planning process for subsequent reasoning under the visual context. This process allows visual information to be written into additional KV cache. This additional cache is visually conditioned yet language-indexed, making it readily transferable to the receiver. A lightweight projector then maps it into the receiver’s decoding prefix and enables effective cross-modal information transfer. Across experimental validations on five multimodal benchmarks, Civil-Relay consistently improves average accuracy over both naive KV-cache projection and text-to-text communication, yielding gains of 7.21 points at 4B and 4.82 points at 8B over naive KV-cache projection.
Our main contributions are as follows:
\begin{enumerate}
  \item We analyze and identify cross-modal attention decay in VLM-to-LLM KV-cache communication, showing that the visual–language KV distribution gap cannot be eliminated by lightweight projection and substantially reduces the receiver's attention to the sender's cross-modal states.

  \item We propose Civil-Relay, which relays a visually conditioned language cache formed during the sender’s planning phase through a lightweight per-layer projector, effectively mitigating cross-modal attention decay.

  \item Experiments across five multimodal benchmarks and two model scales show consistent gains over existing methods, yielding gains of 7.21 points at 4B and 4.82 points at 8B compared with naive KV-cache projection.
\end{enumerate}

\section{Related Work}

\subsection{Multi-Agent Communication}
\label{sec:related_text_latent}

LLM-based multi-agent systems typically coordinate agents through natural-language messages, using role-playing, structured workflows, and multi-agent debate to organize collaboration~\cite{camel23iclr,autogen2024,metagpt2024,chatdev2024,agentverse24iclr,coa2024,debate24icml,chateval23icml,moa2025}. Other studies optimize agent interactions through workflow search, dynamic team composition, and communication topology~\cite{magentic24,gptswarm24,aflow25,dylan24,reconcile24}, or reduce communication overhead by pruning redundant interactions and filtering messages~\cite{agentprune25,agentdropout25,agp26,pact26,messagesieve26}. Beyond text-based communication, recent work explores transferring continuous representations between agents. CIPHER~\cite{cipher2024} communicates through expected word embeddings, while Communicating Activations~\cite{activation2025} and Interlat~\cite{interlat2026} investigate hidden activation exchange and compressed latent states. LatentMAS~\cite{latentmas2026} and HyLaT~\cite{hylat2026} further explore latent working memory and hybrid latent--textual communication. Related work such as Coconut~\cite{coconut2025} studies continuous reasoning within a single model. These approaches reduce reliance on explicit text exchange, while KV-cache communication offers access to intermediate key-value states across layers and token positions.

\subsection{KV Cache Communication}
\label{sec:related_kv}

KV-cache research has extensively explored reducing redundant computation through cache reuse. Prefix caching and inference optimization methods reuse previously computed states~\cite{promptcache24,sglang24,hydragen24}, while other approaches address position offsets, cache scheduling, and interactions between reused segments~\cite{cacheslide26,preble25,cachegen24,cacheblend26,kvlink26,epic26,droidspeak26}. In multi-agent systems, cache reuse extends to shared prompt templates, selective layer sharing, and system-level cache management~\cite{kvreuse2025,kvselect2026,lragent26,tokendance26,cachescout26}. RelayCaching~\cite{relaycaching26} further reuses the Decode KV states of preceding agents to reduce subsequent Prefill computation. Beyond computational reuse, cross-model KV communication transfers internal representations between heterogeneous models. Cache-to-Cache~\cite{c2c2025} projects and fuses KV states from different models under shared inputs, while Dense Latent~\cite{dense2026} transfers Prefill KV states through learned cross-model alignment, including settings where the receiver lacks the original context. Related methods investigate cache fusion~\cite{cachefuse26}, positional transformations and representation alignment~\cite{lcf2026,xkv2026,ucl26}, and calibrated mappings or compressed cache representations~\cite{crossmodelmap26,cachebridge26,heterofold26,cacheback26,condense2026,latcom2026}. Multimodal KV reuse has also been explored within individual models~\cite{vlmcache26}. While these studies address cache reuse, cross-model alignment, and transfer efficiency, our work focuses on the selection of communicated KV states in heterogeneous multimodal systems. Specifically, we investigate visually conditioned planning states as an alternative to directly transferring multimodal Prefill representations.

\section{Setting}
\label{sec:prelim_setting}

We consider a two-agent VLM-to-LLM system in which the two agents share a task but operate on different modalities. The sender \(S\) is a vision-language model equipped with a vision encoder and a text tokenizer. Given an image-question pair \((I,q)\), it compresses its multimodal observation into a message \(m_S\). The receiver \(R\) is a text-only language model that performs downstream reasoning solely from \(m_S\) and a task-specific system prompt \(u\). Formally,
\begin{equation}
    m_S = f_S(I,q), 
    \qquad 
    \hat{y} \sim p_R(\cdot \mid u, m_S),
    \label{eq:setting}
\end{equation}
where \(f_S\) denotes the sender-side encoding, \(p_R\) is the receiver's reasoning distribution, and \(\hat{y}\) is the generated answer. 

Since \(R\) has neither a vision encoder nor visual positional representations, any image information used by the receiver must be conveyed through \(m_S\). This capability decoupling indicates that the sender and receiver maintain different knowledge and representation spaces.

\textbf{Text-to-Text communication.}
Language is a natural information medium and remains the predominant choice in existing multi-agent systems. In this paradigm, the sender verbalizes its multimodal input into a text message \(m_S^{\mathrm{txt}}=(w_1,\ldots,w_T)\), which the receiver re-prefills before generation:
\begin{equation}
    \mathcal{C}^{\mathrm{txt}}_{R}=
    \operatorname{Prefill}_R\!\left(\operatorname{Tokenizer}_R(m_S^{\mathrm{txt}})\right).
    \label{eq:text_comm}
\end{equation}
Although simple and model-agnostic, text-to-text communication suffers from two limitations. First, verbalization inherently loses information, especially for information-dense visual images. The visual details that unable to be spelled out are discarded in transmission. Second, the receiver is forced to re-prefill the relayed text, spending additional computation on content the sender has already encoded.

\textbf{Naive KV cache projection.}
The KV cache encapsulates rich information and intermediate network features ~\cite{c2c2025,dense2026}. Some methods~\cite{dense2026} train a lightweight projection module to pass the sender's prefill KV cache directly to the receiver for reasoning:
\begin{equation}
    \widehat{\mathcal{C}}^{R}=\mathcal{T}_{\phi}\!\left(m_S^{\mathrm{kv}}\right),
\end{equation}
where \(m_S^{\mathrm{kv}}=\mathcal{C}^{S}(I,q)\) denotes the KV cache obtained during the sender's prefill, and \(\mathcal{T}_{\phi}\) denotes the projection performed by a lightweight network \(\phi\). 
Compared with text-to-text communication, naive KV cache projection avoids the additional computation of repeated prefill and delivers hierarchical features directly to the receiver. This form of projection has been shown to enable efficient information transfer between LLMs of similar capability. However, whether it can be transferred across models with different capabilities, such as in the VLM-to-LLM setting, remains an open question.

\begin{figure*}[!t]
  \centering
  \includegraphics[width=\textwidth]{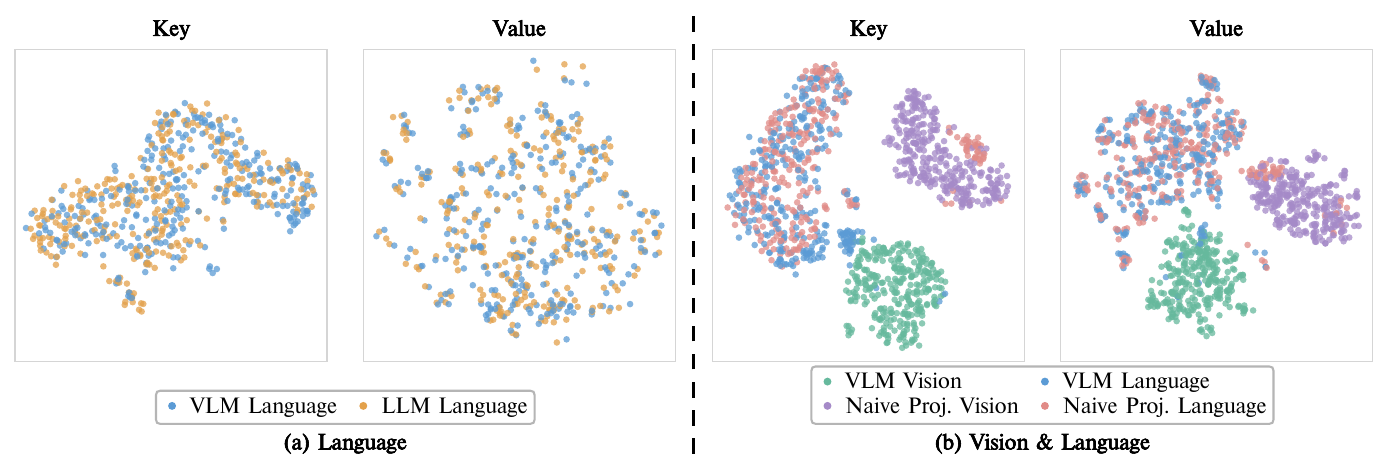}
  \caption{
  t-SNE visualization of KV representations in the 4B setting. We compare token-level representations from the VLM, LLM and Naive Projection using corresponding tokens from MathVision. (a) Language representations exhibit substantial overlap in both Key and Value spaces. (b) Vision representations have distributions markedly different from those of language representations, regardless of whether projection is applied.
  }
  \label{fig:4b_tsne}
\end{figure*}
\begin{table}[t]
\centering
\small
\renewcommand{\arraystretch}{1.05} 
\setlength{\tabcolsep}{4pt}        
\caption{Comparison of VLM-only, Text-to-Text, and Naive KV cache Projection on MMStar and
MathVision for Qwen3 4B and 8B settings. Entries are accuracy (\%). The best and second-best performance are \textbf{bold} and \underline{underlined}.}
\label{tab:4b_ana_cmp}
\begin{tabularx}{\linewidth}{@{} 
  >{\hsize=0.75\hsize\linewidth=\hsize\centering\arraybackslash}X
  >{\centering\arraybackslash}p{2.3cm}
  >{\hsize=1.10\hsize\linewidth=\hsize\centering\arraybackslash}X
  >{\hsize=1.15\hsize\linewidth=\hsize\centering\arraybackslash}X @{}}
\toprule
\multirow{2}{*}{\textbf{Setting}} & \multirow{2}{*}{\textbf{Method}}
& \multicolumn{2}{c}{\textbf{Benchmark}} \\
\cmidrule(lr){3-4}
& & \textbf{MMStar} & \textbf{MathVision} \\
\midrule
\multirow{3}{*}{Qwen3 4B}
& VLM-only         & \textbf{64.67}    & \underline{51.81} \\
& Text-to-Text     & \underline{54.53} & \textbf{52.27}    \\
& Naive Projection & 51.27             & 48.62             \\
\midrule
\multirow{3}{*}{Qwen3 8B}
& VLM-only         & \textbf{66.60}    & \underline{52.43} \\
& Text-to-Text     & 52.87             & \textbf{53.88}    \\
& Naive Projection & \underline{61.27} & 47.80             \\
\bottomrule
\end{tabularx}
\end{table}
\section{Analysis}
\label{sec:analysis}

This section analyzes KV cache communication in the VLM-to-LLM regime. We begin by comparing performance on two multimodal math benchmarks, MMStar~\cite{mmstar2024} and MathVision~\cite{mathvision2024}. Both require the VLM's visual ability to recognize condition information in images as well as the LLM's mathematical reasoning ability. We adopt Qwen3-VL~\cite{qwen3vl2025} as the VLM and Qwen3~\cite{qwen3_2025} as the LLM, where Qwen3 has been reported to possess substantially stronger mathematical capability than Qwen3-VL~\cite{qwen3_2025,qwen3vl2025}. The compared methods include VLM-only (i.e., performance of Qwen3-VL), text-to-text communication, and naive projection, in which the projector is trained to align the layer-wise caches of the VLM and the LLM.

As shown in Table~\ref{tab:4b_ana_cmp}, naive projection underperforms both VLM-only and text-to-text communication, achieving accuracy lower than VLM-only with margins of 13.40 and 5.33 points on MMStar at 4B and 8B, and 3.19 and 4.63 points on MathVision. This gap indicates that transferring the Prefill KV cache is insufficient to recover multimodal inference performance. By contrast, the same KV cache projection has proven effective between LLMs of comparable capability. The failure therefore cannot be attributed to the projection mechanism, but to a mismatch specific to the cross-modal regime.

\begin{figure}[t]
  \centering
  \includegraphics[width=\linewidth]{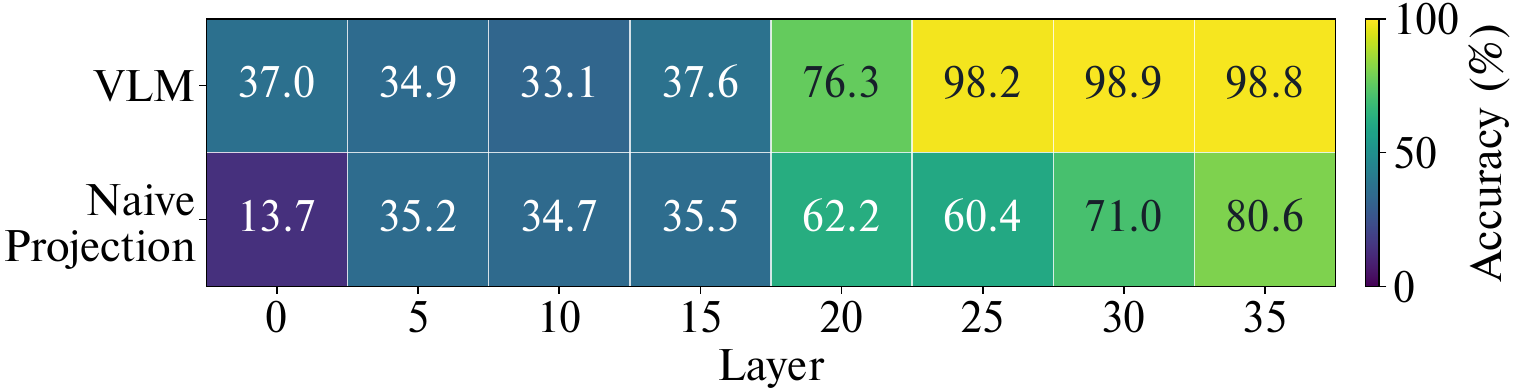}
  \caption{Layer-wise linear probe accuracy on CLEVR. 
  }
  \label{fig:clevr_probe}
\end{figure}
\begin{figure*}[!t] 
    \centering
    \includegraphics[width=1.0\textwidth]{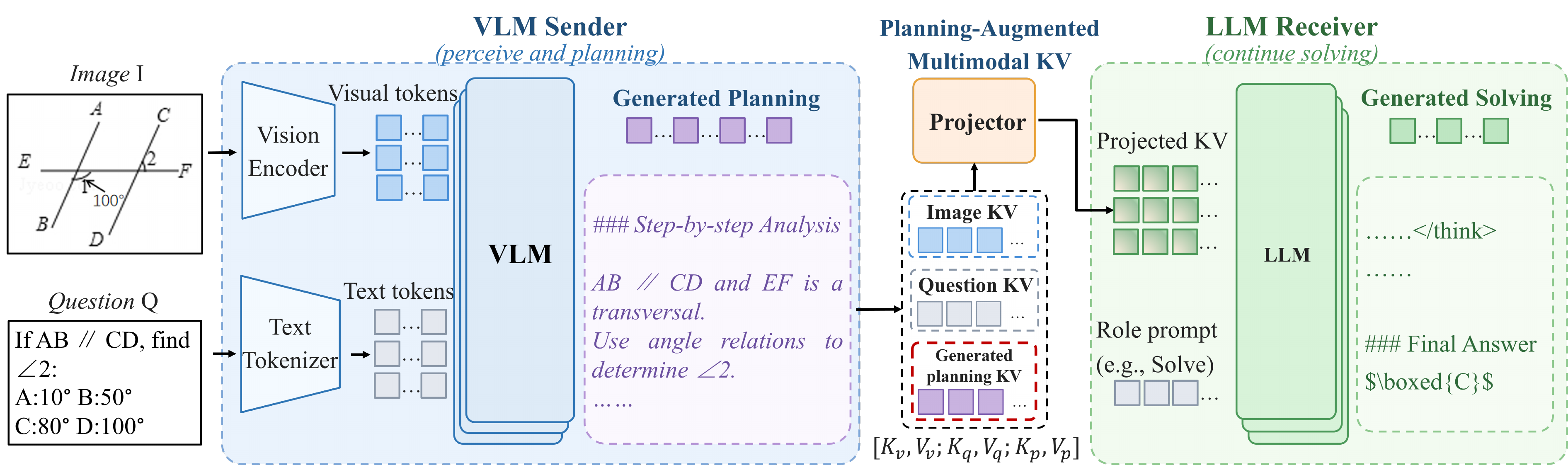}
    \caption{
Overview of Civil-Relay for multimodal multi-agent communication. Given an image and a question, the VLM sender performs multimodal perception and autoregressive task planning, producing KV states associated with the visual input, question, and generated reasoning tokens. A lightweight projector maps the selected sender KV cache into the LLM Receiver's cache
space. The Receiver conditions on the projected cache and its task prompt to autoregressively generate the final answer.}
    \label{fig:framework}
\end{figure*}
\begin{figure}[t]
  \centering
  \includegraphics[width=0.9\linewidth]{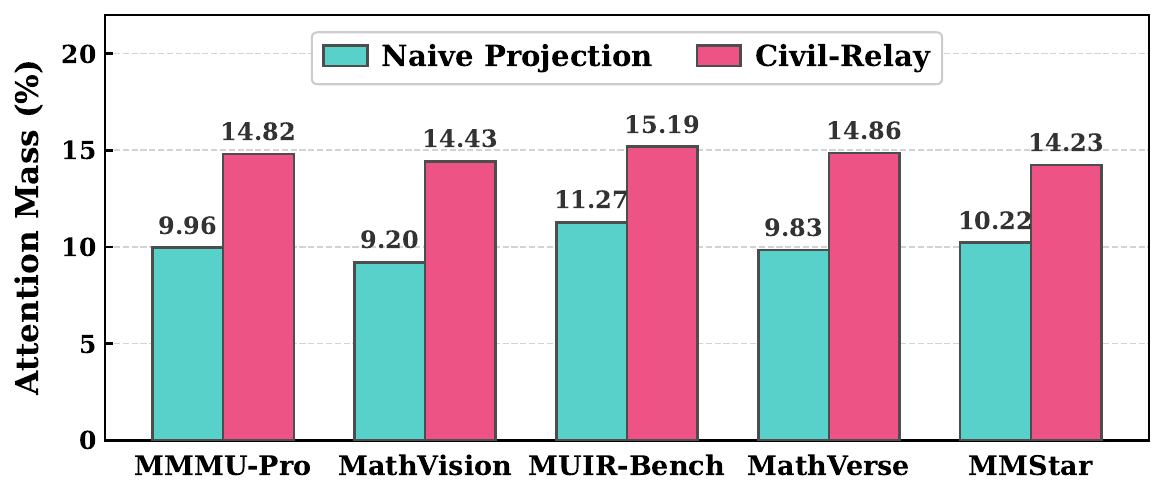}
  \caption{The percentage of the receiver's attention allocated to the sender-related KV cache, averaged over evaluation samples within each benchmark.
  }
  \label{fig:attention_region}
\end{figure}

\subsection{Cross-Modal Distributional Disparity}
\label{sec:analysis_multimodal_structure}

To examine this distinction, we visualize the Key and Value representations of VLM-only and Naive Projection under matched inference trajectories. As shown in Figure~\ref{fig:4b_tsne}(a), language-token features produced by the VLM and LLM exhibit strong alignment, indicating that their language representations follow largely consistent distributions. In contrast, Figure~\ref{fig:4b_tsne}(b) shows that vision-token representations differ substantially from text-token representations in both Key and Value spaces, and this discrepancy persists after projection.


\begin{table}[t]
\centering
\small
\setlength{\tabcolsep}{4pt}
\renewcommand{\arraystretch}{1.05}
\caption{Representation similarity (Linear CKA) across corresponding KV cache components on MathVision in the 4B setting, including similarity between sender-side representations and receiver-side states.}
\label{tab:cka_alignment}
\begin{tabularx}{\linewidth}{
  @{\hspace{4pt}}
  >{\raggedright\arraybackslash}X
  >{\centering\arraybackslash}X
  >{\centering\arraybackslash}X
  @{\hspace{4pt}}
}
\toprule
\textbf{VLM-side State}
& \textbf{vs.\ LLM Text (K)}
& \textbf{vs.\ LLM Text (V)} \\
\midrule
VLM Text     & 0.95 & 0.97 \\
VLM Vision   & 0.34 & 0.59 \\
Proj.\ Vision & 0.31 & 0.56 \\
\bottomrule
\end{tabularx}
\end{table}

We quantify token-wise representational discrepancies using CKA~\cite{kornblith2019}. As reported in Table~\ref{tab:cka_alignment}, the KV caches of text tokens exhibit high structural similarity between the VLM and LLM (95\% for Keys and 97\% Values), whereas vision tokens show considerably lower similarity to text tokens, regardless of whether projection is applied. These results reveal a modality-dependent representational gap, i.e., KV caches of language tokens are largely transferable, while visual tokens occupy distinct subspaces that naive projection cannot fully align.

Beyond the token-level analysis above, we further examine the preservation of visual-modal information at the sequence level on the CLEVR dataset~\cite{johnson2017clevr}. CLEVR requires predicting attributes (e.g., shape and color) from the given image. Using linear probes on the transferred KV caches, we quantify how much vision-relevant information remains after transfer. Figure~\ref{fig:clevr_probe} reports the layer-wise probing results of VLM-only and Naive Projection. As shown, Naive Projection yields significantly lower probing accuracy than VLM-only, achieving 49.16\% average accuracy across all layers compared with 64.36\% for VLM-only. The gap is also evident in deeper layers, where projected features reach 80.6\% while native features achieve 98.2–98.9\%.

\subsection{Cross-Modal Attention Decay}
\label{sec:analysis_language_cache}

The most direct consequence of this cross-modal distributional disparity in the KV cache lies in attention computation. For instance, vision-related keys from the sender may fail to match the receiver’s queries, resulting in low attention scores. We term this phenomenon Cross-Modal Attention Decay. As shown in Figure~\ref{fig:attention_region}, under naive projection, sender-relayed KV states receive only about 10\% of the total attention. This suggests that the poor performance of KV-cache communication on multimodal tasks may stem from insufficient attention to visual information, which in turn impairs downstream reasoning.

The preceding representation analysis points to a promising direction, i.e., increasing the receiver’s attention to the sender’s KV cache may improve downstream performance. Motivated by this, we introduce a simple strategy that guides the sender through a planning process. This planning step encourages the VLM to encode visual information into the KV cache, thereby increasing the receiver’s attention to the encoded visual information. As shown in Figure~\ref{fig:attention_region}, this planning strategy consistently increases the attention across five multi-modal datasets. We will detail this strategy in the next section.

\section{Method}
\label{sec:method}

%

Based on the analysis in Section~\ref{sec:analysis}, we propose Conditioned-on Visual Language Cache Relay (Civil-Relay) for KV cache communication in MAS.
As illustrated in Figure~\ref{fig:framework}, Civil-Relay decomposes communication into three coordinated components: 
\begin{itemize}
    \item A VLM-based sender $S$ that perceives multimodal inputs and encodes them into a KV cache;
    \item A layer-wise projector network $\mathcal{P}$ that transfers the KV cache from the sender to the receiver distribution;
    \item An LLM-based receiver $R$ that performs downstream reasoning and task solving based on the projected KV cache.
\end{itemize}
Given an image \(I\) and a question \(Q\), the pipeline proceeds through sender prefilling, KV cache projection, and receiver reasoning are detailed as follows.

\paragraph{Sender prefilling.}
The VLM sender \(S\) first prefills the visual and text tokens and then plans for downstream reasoning, producing a plan grounded in the visual context.
During planning, the sender inspects the image and question to determine which visual evidence the task requires, and records this evidence together with the associated numerical cues in the plan.
Formally, let \(\mathrm{Prompt}_{\mathrm{plan}}\) denote the system prompt for planning, the sender prefilling process obtains the KV cache as
\begin{equation}
    \mathcal C^S(I,q)
    = \operatorname{Prefill}_S\!\left(
    \mathrm{Prompt}_{\mathrm{plan}}, I, q \right).
    \label{eq:sender_prefill}
\end{equation}
Here, \(\mathcal C^S(I,q)\) contains the layer-wise Keys and Values for the planning prompt, visual tokens, and question tokens. Input encoding is included in \(\operatorname{Prefill}_S\),
and the fixed planning prompt is omitted from the cache arguments
for brevity. The sender then generates \(z\) autoregressively from this context,
extending the cache to \(\mathcal C^S(I,q,z)\).

\paragraph{KV cache projection.}
The projector bridges the representation gap between the two backbones, mapping the relayed sender cache into the receiver's KV space so that the receiver can condition on it without adaptation.
Crucially, we make no assumption that the sender and receiver share the same KV-head count or head dimension.
To this end, we flatten each cache entry along the head axis and adapt the dimension of the KV cache through the separate two-layer MLPs projections \(\mathcal P^K\) and \(\mathcal P^V\) for Keys $K^S$ and Values $V^S$.

A remaining obstacle is the mismatch in positional geometry, i.e., the sender places visual tokens on multi-axis M-RoPE positions over time, height, and width, whereas a text-only receiver uses one-dimensional RoPE.
Projecting Keys directly would entangle features with two different position geometries.
We resolve this by de-rotating Keys before projection and re-rotating them afterwards.
For each relayed token \(t\), let \(K_t^S\) and \(V_t^S\) denote its cached Key and Value at the matched sender layer, respectively. Let \(\mathcal R_S(t)\) denote its original sender rotation and \(\mathcal R_R(t)\) the rotation at its assigned receiver position. These operators act head-wise and may use different position coordinates. With \(\widetilde K_t^S=\mathcal R_S(t)^\top K_t^S\) denoting the de-rotated Key, the conversion is
\begin{equation}
\begin{aligned}
    \widehat K_t^R &= \mathcal R_R(t)\,\mathcal P^K(\widetilde K_t^S),\\
    \widehat V_t^R &= \mathcal P^V(V_t^S).
\end{aligned}
\label{eq:rope_conversion}
\end{equation}

For each receiver layer, we collect the converted Keys and Values over the relayed tokens in their assigned receiver order. These per-layer KV pairs together form the projected cache \(\widehat{\mathcal C}^{R}\):
\begin{equation}
    \widehat{\mathcal C}^{R}
    = \mathcal T_\phi\!\left(
    \mathcal C^S(I,q,z)\right)
    =[K_v,V_v;K_q;V_q;K_p;V_p].
    \label{eq:cache_transfer}
\end{equation}
Here, \(\mathcal T_\phi\) comprises layer matching, Key de-rotation, KV projection, and Key re-rotation as described above. $(K_v,V_v)$, $(K_q,V_q)$, and $(K_p,V_p)$
denote the key and value caches of image,
question, and planning tokens, respectively.

\paragraph{Receiver reasoning.}
With the cache now expressed in the receiver's KV space, the LLM receiver \(R\) conditions on it as a prefix and performs downstream reasoning and task solving alongside its task prompt \(u\).
\begin{equation}
    \hat y_t \sim p_R\!\left(
    \cdot \mid u,\hat y_{<t},\widehat{\mathcal C}^{R}\right).
    \label{eq:receiver_decoding}
\end{equation}

\subsection{Projector Training}
\label{sec:method_training}

In Civil-Relay, the sender and receiver remain frozen, and only the projector \(\phi\) is trained to bridge the two KV spaces.
The training process consists of two stages. Stage~I supervises the projector to reproduce the receiver's native KV features on token-aligned contexts, while Stage~II shifts supervision to the receiver's output distribution once the projected cache is injected.

\paragraph{Stage I: KV feature alignment.}
This stage targets the representational role of the projector, teaching it to map sender features onto the KV statistics that the receiver expects.
Keys on both sides are de-rotated using their respective native positions, denoted as \(\widetilde K^S\) and \(\widetilde K^R\). The objective is
\begin{equation}
\begin{aligned}
    \mathcal L_{\mathrm{MSE}} = \frac12\,[
    \operatorname{MSE}\!\left(\mathcal P^K(\widetilde K^S),\widetilde K^R\right)+\operatorname{MSE}\!\left(\mathcal P^V(V^S),V^R\right)\big].
\end{aligned}
\label{eq:feature_mse}
\end{equation}
This warm-up stage establishes a preliminary alignment between the sender and receiver KV spaces and provides a stable initialization for the subsequent behavior alignment.

\paragraph{Stage II: Receiver behavior alignment.}
Beyond token-wise alignment, the second stage supervises the projector through the receiver's output distribution.
Specifically, for each sample, we obtain the Teacher distribution by forwarding the receiver without the sender cache, and the Student distribution by forwarding the receiver with the injected projected cache. Both read logits at the prediction positions of the same reference sequence \(y\).
Let \(p^T\) and \(p^\phi\) represent the Teacher and Student logits for sample \(m\) at position \(t\),
\begin{equation}
    \mathcal L_{\mathrm{KL}}
    =\tau^2
    D_{\mathrm{KL}}\!\left(p^T\,\|\,p^\phi\right)
    \label{eq:behavior_kl}
\end{equation}
where \(\tau\) is the distillation temperature.
Notably, Teacher logits are detached during training, ensuring backward signal reaches the projector only through the Student's decoding graph.
\section{Experiment}

\begin{table*}[t]
\centering
\caption{
Performance (accuracy, \%) of different methods in the 4B and 8B settings. The column with ``Average'' indicates the unweighted mean across five benchmarks.
For each model size, the best and second-best results in each column
are \textbf{bold} and \underline{underlined}, respectively.
}
\label{tab:main_results}
\resizebox{0.9\textwidth}{!}{
\renewcommand{\arraystretch}{1.3}
\begin{tabular}{c l | c c | c c c  | c}
\toprule
\multirow{2}{*}{\textbf{Model}}
& \multirow{2}{*}{\textbf{Method}}
& \multicolumn{2}{c|}{\textbf{Mathematical Reasoning}}
& \multicolumn{3}{c|}{\textbf{General Multimodal Understanding}}
& \multirow{2}{*}{\textbf{Average}} \\
\cmidrule(lr){3-4}\cmidrule(lr){5-7}
& & MathVerse & MathVision & MUIR-Bench & MMMU-Pro & MMStar & \\
\midrule
\multirow{5}{*}{4B}
& VLM-only
& 57.28 & 51.81 & \textbf{58.73} & 55.14
& \underline{64.67} & \underline{57.53} \\
& Text-to-Text
& \underline{59.37} & \underline{52.27} & 57.12 & 53.82
& 54.53 & 55.42 \\
& C2C
& 1.93 & 4.14 & 16.92 & 13.70
& 16.13 & 10.57 \\
& Naive Projection
& 59.09 & 48.62 & 47.54 & \underline{60.23}
& 51.27 & 53.35 \\
& \textbf{Ours}
& \textbf{63.40} & \textbf{53.16} & \underline{58.15}
& \textbf{62.89} & \textbf{65.20} & \textbf{60.56} \\
\midrule
\multirow{5}{*}{8B}
& VLM-only
& \underline{58.65} & \underline{52.43} & \underline{57.00}
& 58.79 & \underline{66.60} & \underline{58.69} \\
& Text-to-Text
& 55.36 & \textbf{53.88} & 51.69 & 48.27
& 52.87 & 52.41 \\
& C2C
& 3.20 & 2.24 & 15.88 & 12.31
& 6.93 & 8.11 \\
& Naive Projection
& 55.94 & 47.80 & 50.81 & \underline{60.17}
& 61.27 & 55.20 \\
& \textbf{Ours}
& \textbf{58.76} & 49.70 & \textbf{59.65}
& \textbf{63.64} & \textbf{68.33} & \textbf{60.02} \\
\bottomrule
\end{tabular}}
\end{table*}

\subsection{Experimental Setup}
\label{sec:exp_setup}

\paragraph{Implementation details.}
We use Qwen3-VL~\cite{qwen3vl2025} as the sender and Qwen3~\cite{qwen3_2025} as the receiver. At both the 4B and the 8B scale, sender and receiver have the same size. Training updates only the projector between the LLM of the sender VLM and the receiver LLM, while the vision encoder and both pretrained language models stay frozen. The projector is a two-layer MLP without residual connections, with hidden dimension 1,024 and SiLU activation. Training uses 500K text-only samples from OpenHermes-2.5 with a maximum sequence length of 2,048. Both stages use AdamW with a learning rate of $2\times10^{-5}$ and a global batch size of 32, and the KL stage uses a learning-rate warmup ratio of 0.1. All experiments run on 8$\times$ Huawei Ascend 910B NPUs (64GB). At inference, sender and receiver keep their default sampling parameters, with context lengths of 16,384 for the sender and 32,768 for the receiver.

\paragraph{Benchmarks.}
We evaluate the proposed method on five multimodal benchmarks covering visual mathematical reasoning, cross-disciplinary multimodal understanding, visual perception and reasoning, and multi-image understanding. \textbf{MathVerse}~\cite{mathverse2024} targets visual mathematical reasoning over charts and geometric diagrams and probes visual understanding through different forms of information presentation; \textbf{MathVision}~\cite{mathvision2024} collects visual problems from real mathematics competitions and spans 16 mathematical domains and 5 difficulty levels; \textbf{MMMU-Pro}~\cite{mmmupro2025} targets cross-disciplinary multimodal academic questions and evaluates professional knowledge understanding and reasoning from images and text; \textbf{MMStar}~\cite{mmstar2024} emphasizes the necessity of visual information and covers six abilities, namely coarse-grained perception, fine-grained perception, instance reasoning, logical reasoning, mathematics, and science and technology; \textbf{MUIR-Bench}~\cite{muirbench2025} targets multi-image understanding and evaluates how a model comprehends multiple images and their relations. Both model scales are evaluated by answer accuracy, which serves as the primary metric.

\paragraph{Communication settings.}
We compare four agent communication settings. \textbf{VLM-only} performs multimodal reasoning directly with the vision-language model and involves no cross-agent context transfer; \textbf{Text-to-Text} communicates the text generated by the sender, which the receiver re-prefills before decoding; \textbf{Naive Projection} shares the projector and training configuration of Civil-Relay but communicates the cache the sender forms at multimodal Prefill; \textbf{Civil-Relay} is our method, which relays the Prefill cache together with the decode-stage planning cache, both mapped through the KV Transmission Projector. All four use the same sender, receiver, and evaluation pipeline.

\subsection{Main Results}
\label{sec:exp_main}

Table~\ref{tab:main_results} summarizes the results on the five multimodal benchmarks. At both the 4B and the 8B scale, Civil-Relay achieves the best accuracy on four of the five benchmarks, which indicates that its advantage does not hinge on a particular model scale or task type. Averaging the accuracy over the five benchmarks, Civil-Relay reaches 60.56\% at 4B and 60.02\% at 8B, improving over VLM-only by 3.03 and 1.32 points, over Text-to-Text by 5.14 and 7.60 points, and over Naive Projection by 7.21 and 4.82 points, respectively.

Across all ten model-scale--benchmark combinations, Civil-Relay outperforms Naive Projection, which points to the communicated content itself. Handing the entire multimodal cache formed at Prefill to the receiver and relaying only the planning cache that the sender organizes under visual conditioning provide information of different structures. Compared with VLM-only, Civil-Relay again improves on four of the five benchmarks at both scales, and is slightly lower only on MUIR-Bench (58.15\% vs.\ 58.73\% in 4B) and MathVision (49.70\% vs.\ 52.43\% in 8B). Overall, Civil-Relay consistently improves communication between heterogeneous multimodal agents.

\subsection{Further Discussion}
\label{sec:exp_discussion}

\paragraph{Is a visually conditioned planning cache sufficient for multimodal reasoning?}
The main experiments show that Naive Projection, which relays the entire cache formed at Prefill, is overall weaker than Civil-Relay. This raises a more specific question, namely whether the receiver needs explicit access to the full visual cache or can obtain what it needs mainly from the planning cache that the sender generates under visual conditioning. To study this, we control the cache region visible to the receiver and construct four settings: question only (Q), vision and question (V+Q), question and planning (Q+P), and vision, question, and planning (V+Q+P). Each setting is a different choice of the relayed set $\mathcal{I}_{\mathrm{relay}}$ of \S\ref{sec:method}. Positions in P are produced by the sender \emph{after} it reads the image and the question, so they are language tokens that carry visual context rather than an ordinary text cache.

\begin{table}[!t]
\centering
\small
\renewcommand{\arraystretch}{1.15}
\caption{Ablation on the receiver-readable KV region across MathVerse, MMStar, and MMMU-Pro. Q, V, and P denote question, visual, and planning positions, respectively, and a checkmark marks a region visible to the receiver.}
\label{tab:6_vis_pilot}
\begin{tabular*}{\linewidth}{
  @{\hspace{6pt}\extracolsep{\fill}}
  c c c c c c
  @{\hspace{6pt}}
}
\toprule
\multirow{2}{*}{\textbf{V}} & \multirow{2}{*}{\textbf{Q}} & \multirow{2}{*}{\textbf{P}}
& \multicolumn{3}{c}{\textbf{Accuracy (\%)}} \\
\cmidrule(lr){4-6}
& & & \textbf{MathVerse} & \textbf{MMStar} & \textbf{MMMU-Pro} \\
\midrule
               & \(\checkmark\) &                & 32.99 & 41.40 & 45.49 \\
\(\checkmark\) & \(\checkmark\) &                & 58.86 & 54.27 & 59.42 \\
               & \(\checkmark\) & \(\checkmark\) & 62.03 & 63.93 & 61.27 \\
\(\checkmark\) & \(\checkmark\) & \(\checkmark\) & \textbf{63.30} & \textbf{67.40} & \textbf{62.95} \\
\bottomrule
\end{tabular*}
\end{table}
As shown in Table~\ref{tab:6_vis_pilot}, the visual cache is nearly dispensable. Relaying only the question and the planning cache (Q+P) comes within 1.27, 3.47, and 1.68 points of the fully visible setting on MathVerse, MMStar, and MMMU-Pro. The planning cache is not merely a substitute, however, as adding it on top of the visual cache improves V+Q by 4.44, 13.13, and 3.53 points. The two are therefore complementary, and planning supplies the larger marginal gain. From an information-transfer perspective, the planning cache keeps the sender's joint processing of image and question in a language representation that a successor agent can read directly, shifting cross-agent communication from a hand-over of raw visual features toward task-oriented information organization.

\paragraph{Retention length of the planning cache.}
Having compared different visible regions, we further ask how the transfer effect of the planning cache depends on how much of it is retained. We retain the first $\rho\in\{0,25,50,75,100\}\%$ of the planning positions in their generation order, and record the corresponding average number of retained tokens together with the accuracy of the receiver (see Table~\ref{tab:4b_ana_ret_plan}).

Full retention gives the highest accuracy, 62.13\%, which is 1.86 points above the $\rho=0$ condition. As the retention ratio grows from 25\% to 75\%, accuracy reaches 58.07\%, 58.07\%, and 58.33\%, respectively. These three values are close to each other and all remain below full retention. The benefit therefore does not grow linearly with the retention ratio but jumps once the planning cache is kept complete. A complete planning context provides more coherent support for reasoning, whereas partial retention may break the organization and continuity of the information. Together with the visibility experiment, this indicates that \emph{The effectiveness of KV cache communication depends on both the visual context encoded in the cache and how the sender's planning information is organized within it}.

\begin{table}[!t]
\centering
\small
\caption{Accuracy of the receiver under different planning-cache retention ratios.}
\label{tab:4b_ana_ret_plan}
\begin{tabularx}{\linewidth}{@{} >{\centering\arraybackslash}X
                                 c
                                 >{\centering\arraybackslash}X @{}}
\toprule
Retention (\%) & Avg.\ Retained Tokens & Accuracy (\%)\\
\midrule
0   & 0.0   & 60.27\\
25  & 90.9  & 58.07\\
50  & 182.4 & 58.07\\
75  & 273.6 & 58.33\\
100 & 365.2 & 62.13\\
\bottomrule
\end{tabularx}
\end{table}
\begin{table}[!t]
\centering
\small
\setlength{\tabcolsep}{4pt}
\renewcommand{\arraystretch}{1.15}
\caption{Effect of planning cache on MMStar. }
\label{tab:6_cache_effect}
\begin{tabular*}{\linewidth}{
  @{\hspace{6pt}\extracolsep{\fill}}
  l c c
  @{\hspace{6pt}}
}
\toprule
\textbf{Condition} & \textbf{Accuracy (\%)} & \textbf{$\Delta$ vs. Correct} \\
\midrule
Correct Image KV & \textbf{65.00} & --      \\
Donor Image KV   & 59.87          & $-5.13$ \\
Grey Image KV    & 59.87          & $-5.13$ \\
\bottomrule
\end{tabular*}
\end{table}

\paragraph{Does the planning cache carry visual context beyond the surface text?}
The preceding results demonstrate the high utility of Planning KV. One question nevertheless remains to be settled, namely whether the KV representation genuinely preserves visual context or merely acts as a redundant encoding of the surface text. To answer this, we conduct controlled ablations by varying the visual source during cache construction, as shown in Table~\ref{tab:6_cache_effect}. We compare the effect of different visual sources fed into our method. When the receiver is given the KV constructed from the correct visual context, Ours achieves a peak accuracy of 65.00\%. In contrast, substituting the visual source with a donor image or a grey image immediately degrades the performance to 59.87\%. This significant drop provides direct evidence that the planning cache is highly sensitive to the visual input, successfully encapsulating visual evidence beyond what the explicit planning text can convey.

\section{Conclusion}

In this paper, we study KV-cache communication in the VLM-to-LLM regime. Our analysis shows that visual and language KV caches exhibit substantial distributional divergence, and that this modality-specific gap cannot be eliminated by lightweight projection. Therefore, the receiver allocates insufficient attention to sender-relayed visual states, a phenomenon we term cross-modal attention decay. To address this issue, we propose Civil-Relay, which relays a visually conditioned language cache formed during the sender's planning phase through a lightweight per-layer projector into the receiver's decoding prefix. On five multimodal benchmarks, Civil-Relay improves average accuracy over naive KV-cache projection by 7.21 points at 4B and 4.82 points at 8B, while also surpassing text-to-text communication at both scales.



{
    \small
    \bibliographystyle{ieeenat_fullname}
    \bibliography{main}
}

\clearpage
\setcounter{page}{1}
\appendix
\twocolumn[
    \begin{center}
        \LARGE\bfseries Appendix
    \end{center}
    \vspace{0.5em}
]


%

\section{Implementation Details}
We train the Cache Projector on the OpenHermes-2.5 dataset~\cite{openhermes2023} using a two-stage pipeline consisting of plain MSE alignment followed by KL distillation. Each stage is trained for one epoch with both the Sender and Receiver kept frozen and only the Projector parameters are updated. To maximize evaluation throughput, we adopt vLLM~\cite{kwon2023vllm} as the inference backend and support Huawei Ascend NPUs via vLLM-Ascend. While VLM-only and Text-to-Text baselines follow standard inference paths, for Naive Projection~\cite{dense2026} and Civil-Relay we extend the KV Cache export and injection pathways in vLLM, allowing the Sender's KV Cache to be mapped through the frozen Projector and injected as prefix cache for the Receiver's generation. Specifically, Keys undergo positional encoding transformation in the order of deRoPE-Projector-RoPE, whereas Values are projected directly. Both the Sender and Receiver use their respective model-default sampling parameters, with a repetition penalty of 1.1 and a presence penalty of 1.5. For evaluation, we employ asynchronous requests, data parallelism, and continuous batching to enhance throughput, while reusing the same generated Planning token sequences to ensure a consistent and fair comparison between Text-to-Text and Civil-Relay.

\subsection{Prompt Settings}
As illustrated in Figure~\ref{fig:supp_prompts}, rather than imposing overly elaborate instruction prompts to force the VLM and LLM to follow rigid rules for planning and reasoning during inference, we employ concise prompts that instruct the VLM to carefully inspect both the image and the question. Specifically, the planning stage is prompted to incorporate essential visual evidence and relevant quantitative details, while clearly articulating key visual cues so that the Receiver can integrate textual descriptions with the visual input to guide downstream reasoning. The LLM then builds upon the VLM’s visual grounding and planning trajectory to solve the tasks. Experimental results further demonstrate that such lightweight prompts effectively elicit Conditioned-on-Vision Language Caches in the VLM, enabling the LLM to seamlessly access information from the Sender and accomplish subsequent reasoning tasks.
\begin{figure}[t]
    \centering

    \begin{promptbox}{PlannerBlue}{Sender Prompt}
        You are the Planner in a two-agent system.
        Examine the image and problem carefully.
        Develop a concise, reliable solution plan that
        includes the visual evidence, relevant quantities,
        and reasoning needed to solve the task.
        Do not assume the Solver can inspect the image.
    \end{promptbox}

    \begin{promptbox}{SolverGreen}{Receiver Prompt}
        You are the Solver in a two-agent system.
        Use the Planner's transmitted reasoning together
        with your own reasoning to answer the task.
        Return the final answer clearly and follow the
        task's required answer format.
    \end{promptbox}

    \caption{
        Role-specific prompts for the Sender and Receiver in Civil-Relay.
    }
    \label{fig:supp_prompts}
\end{figure}

\subsection{Benchmark Details}
To evaluate heterogeneous agent collaboration across diverse multimodal scenarios, we select five representative benchmarks spanning two core categories: \textbf{multimodal mathematical reasoning} and \textbf{general multimodal understanding and reasoning}.

\begin{table*}[!t]
\centering
\caption{
Performance (accuracy, \%) across five benchmarks under
different Sender--Receiver configurations.
``--'' denotes single-model inference without a separate
receiver. Average is the unweighted mean across five benchmarks.
Within each communication configuration, the best and
second-best results among communication methods are
\textbf{bold} and \underline{underlined}, respectively.
Tied best results are both bolded.
}
\label{tab:suppl_results}

\begingroup
\renewcommand{\arraystretch}{1.18}
\setlength{\tabcolsep}{5pt}

\resizebox{0.96\textwidth}{!}{%
\begin{tabular}{c c l | c c | c c c | c}
\toprule
\multirow{2}{*}{\textbf{Sender}}
& \multirow{2}{*}{\textbf{Receiver}}
& \multirow{2}{*}{\textbf{Method}}
& \multicolumn{2}{c|}{\textbf{Mathematical Reasoning}}
& \multicolumn{3}{c|}{\textbf{General Multimodal Understanding}}
& \multirow{2}{*}{\textbf{Average}} \\
\cmidrule(lr){4-5}
\cmidrule(lr){6-8}
& & & MathVerse & MathVision
& MUIR-Bench & MMMU-Pro & MMStar & \\
\midrule


\multirow{4}{*}{Qwen3-VL 4B}
& --
& VLM-only
& 57.28 & 51.81 & 58.73 & 55.14
& 64.67 & 57.53 \\

\cmidrule(lr){2-9}

& \multirow{3}{*}{Qwen3 8B}
& Text-to-Text
& \underline{56.90}
& \textbf{49.41}
& \textbf{61.92}
& \underline{56.88}
& \textbf{67.20}
& \textbf{58.46} \\

& & Naive Projection
& 55.66
& \underline{46.12}
& 44.58
& 56.42
& 56.47
& 51.85 \\

& & \textbf{Ours}
& \textbf{57.31}
& 45.49
& \underline{58.35}
& \textbf{58.73}
& \textbf{67.20}
& \underline{57.42} \\

\midrule


\multirow{4}{*}{Qwen3-VL 8B}
& --
& VLM-only
& 58.65 & 52.43 & 57.00 & 58.79
& 66.60 & 58.69 \\

\cmidrule(lr){2-9}

& \multirow{3}{*}{Qwen3 4B}
& Text-to-Text
& \underline{58.07}
& \textbf{51.61}
& \textbf{58.88}
& \underline{61.73}
& \underline{66.87}
& \underline{59.43} \\

& & Naive Projection
& 57.87
& 44.51
& 50.38
& 60.35
& 57.40
& 54.10 \\

& & \textbf{Ours}
& \textbf{58.68}
& \underline{49.97}
& \underline{58.85}
& \textbf{63.29}
& \textbf{69.73}
& \textbf{60.10} \\

\bottomrule
\end{tabular}%
}
\endgroup
\end{table*}

\paragraph{Multimodal Mathematical Reasoning}
We choose \textbf{MathVerse}~\cite{mathverse2024} and \textbf{MathVision}~\cite{mathvision2024} to assess the synergy between visual perception and multi-step symbolic reasoning. These tasks require extracting geometric structures and quantitative relationships from images and integrating them with textual conditions for complex deduction. In our VLM-to-LLM setting, the VLM handles visual perception and initial task planning, while the LLM executes downstream mathematical derivation, thereby testing the complementary strengths of both models.

\begin{itemize}
    \item \textbf{MathVerse:} Comprising 2,612 visual math problems across domains such as geometry and functions, MathVerse constructs multiple problem versions with varying ratios of visual-to-textual information. This design explicitly probes a model's genuine reliance on visual inputs and quantifies how visual context impacts mathematical reasoning.
    \item \textbf{MathVision:} Containing 3,040 competition-level multimodal problems across 16 mathematical disciplines and multiple difficulty tiers, MathVision places greater emphasis on deep, multi-step mathematical deduction, requiring tight integration of fine-grained visual cues with domain-specific mathematical knowledge.
\end{itemize}

\paragraph{Multimodal Understanding and Reasoning}
We select \textbf{MMStar}~\cite{mmstar2024}, \textbf{MMMU-Pro}~\cite{mmmupro2025}, and \textbf{MUIR-Bench}~\cite{muirbench2025} to evaluate cross-modal information transfer across broader visual domains, including natural scenes, specialized academic diagrams, and multi-image contexts. These benchmarks impose diverse demands on fine-grained visual extraction, cross-modal integration, and knowledge-intensive reasoning, verifying whether the VLM's vision-conditioned cache provides sufficient context for the LLM to solve downstream tasks.

\begin{itemize}
    \item \textbf{MMStar:} Consisting of 1,500 carefully curated samples spanning six core capabilities—coarse perception, fine-grained perception, instance reasoning, logical reasoning, mathematics, and science \& technology—MMStar enforces strict visual dependency by filtering out samples solvable via text-only shortcuts.
    \item \textbf{MMMU-Pro:} Built to evaluate college-level subject knowledge and complex reasoning across science, engineering, medicine, and the arts, MMMU-Pro increases both task difficulty and visual dependency compared to the original MMMU, rigorously testing joint reasoning over specialized domain knowledge and visual evidence.
    \item \textbf{MUIR-Bench:} A comprehensive multi-image understanding benchmark containing 2,600 questions across tasks such as image ordering, visual grounding, and cross-image relation reasoning. Unlike single-image benchmarks, it requires aggregating information across multiple images and capturing their spatial, temporal, or semantic dependencies.
\end{itemize}

\section{Cross-Scale Benchmark Results}

The main paper evaluation establishes cache communication between a VLM Sender and an LLM Receiver at matching model scales. We extend this setting to 4B$\rightarrow$8B and 8B$\rightarrow$4B to examine whether the proposed alignment remains effective when Sender and Receiver capacities differ. Evaluating both directions distinguishes transfer to a larger Receiver from transfer to a smaller Receiver, which impose different cache-mapping requirements.

\begin{figure*}[!t]
  \centering
  \includegraphics[width=\textwidth]{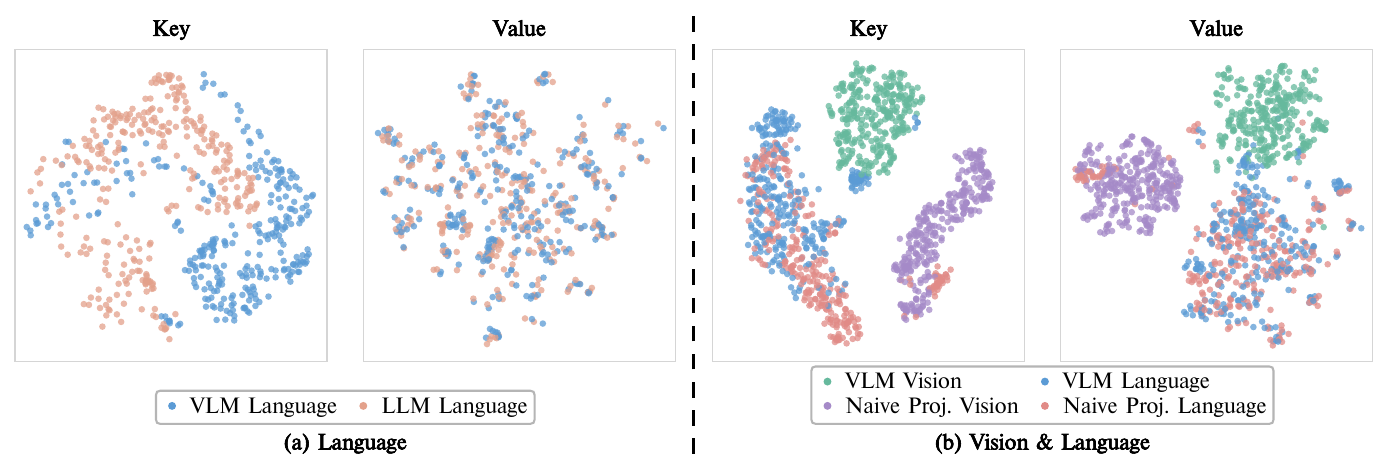}
  \caption{
  t-SNE visualization of KV representations in the 8B setting. We compare token-level representations from the VLM, LLM and Naive Projection using corresponding tokens from MathVision. (a) Language representations exhibit substantial overlap in the Value space, while retaining partially separated local structures in the Key space. (b) Vision representations have distributions markedly different from those of language representations, regardless of whether projection is applied.
  }
  \label{fig:8b_tsne}
\end{figure*}
\begin{figure*}[!t]
  \centering
  \includegraphics[width=\textwidth]{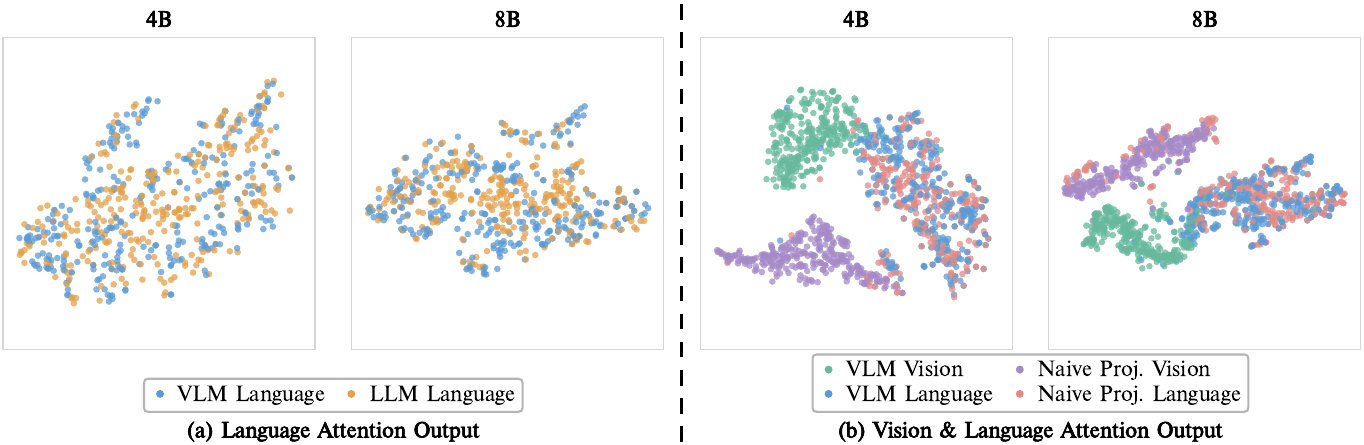}
  \caption{
  t-SNE visualization of Attention Output representations in the 4B and 8B setting. We compare token-level representations from the VLM, LLM and Naive Projection using corresponding tokens from MathVision. (a) Language representations exhibit substantial overlap in Attention Output spaces. (b) Vision representations have distributions markedly different from those of language representations, regardless of whether projection is applied.
  }
  \label{fig:4b8b_attn_tsne}
\end{figure*}
As shown in Table~\ref{tab:suppl_results}, for each direction we train a projector using the same two-stage objective as in the main paper, with the layer mapping and cache dimensions adapted to the selected model pair. We compare communication methods using the same Sender and Receiver checkpoints, benchmark items, role prompts, decoding configuration, and scorer within that direction. Across the two transfer directions, Civil-Relay outperforms Naive Projection in nine of ten benchmark configurations, achieving the highest overall average in the 8B$\rightarrow$4B setting while remaining competitive with a lower average than Text-to-Text in the 4B$\rightarrow$8B setting. These results confirm that our vision-conditioned language cache remains effective across both matched and mismatched model scales. 

We also observe two notable behaviors across directions. First, in the 4B$\rightarrow$8B MathVision setting, where Civil-Relay trails Naive Projection, we observe occasional repetitive token generation that prevents task completion, suggesting that residual cache mismatch across transferred Planning states can disrupt Receiver decoding. Second, the directional asymmetry suggests that cache transfer depends on the relative capacities of the Sender and Receiver. In the 4B$\rightarrow$8B setting, Text-to-Text allows the larger Receiver to re-encode the smaller Sender's Planning using its own higher-capacity network, whereas direct cache transfer requires it to operate on projected states formed by the smaller model. In the reverse direction, the larger Sender can provide richer visually conditioned representations to the smaller Receiver. This interpretation explains why the advantage of Civil-Relay over Text-to-Text is more pronounced in the 8B$\rightarrow$4B setting.

\begin{figure*}[!t]
  \centering
  \includegraphics[width=\textwidth]{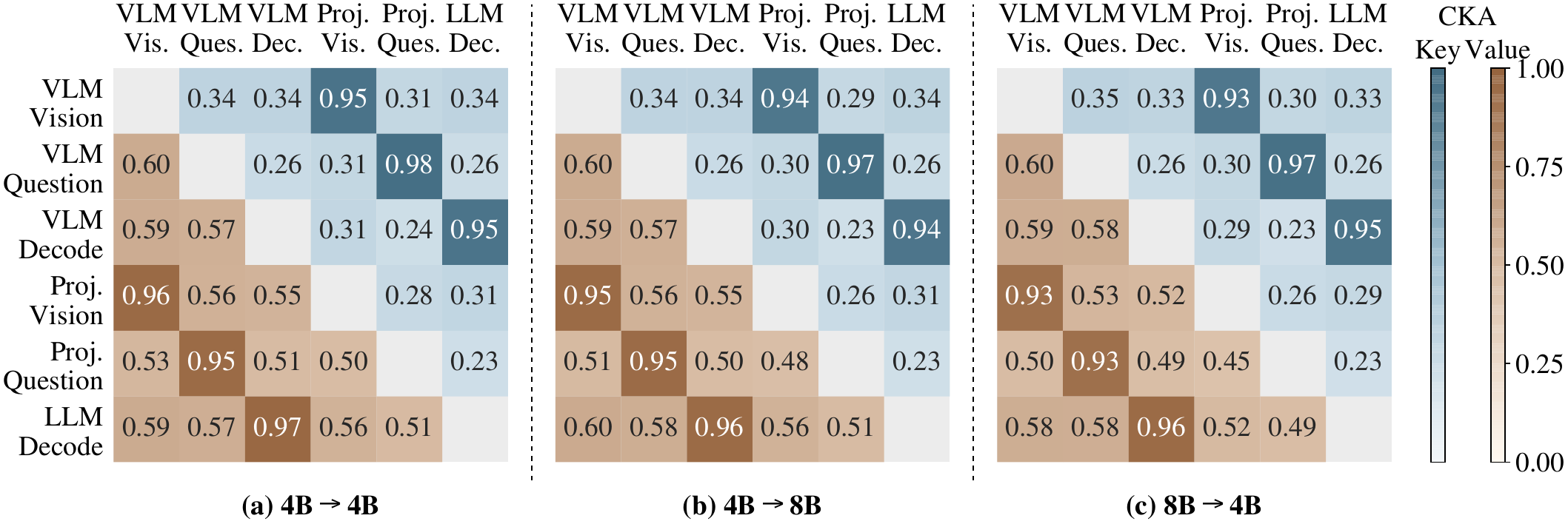}
  \caption{
  Representation similarity across KV-cache components for (a) 4B$\rightarrow$4B, (b) 4B$\rightarrow$8B, and (c) 8B$\rightarrow$4B communication. Each panel reports linear CKA among MLLM vision, question, and decoding representations, projected vision and question representations, and LLM decoding representations, averaged across layers. The upper triangle (blue) shows Key similarity, and the lower triangle (orange) shows Value similarity; both share the same color scale. Gray diagonal entries indicate self-similarity (CKA = 1) and are omitted. Corresponding representation groups exhibit consistently high similarity across both same-scale and cross-scale communication.
  }
  \label{fig:4b8b_cka}
\end{figure*}

\section{Additional Representation Results}
\label{sec:supp_representation}

The main paper t-SNE~\cite{vandermaaten2008} visualization and CKA~\cite{kornblith2019} summary motivate a closer examination of how visual and language cache representations behave under projection. We extend the analysis to examine whether visual-language separation persists at the 8B scale, compare the structural similarity of native and projected representations using CKA, and evaluate how task-relevant visual information can be decoded across layers using linear probes. Together, these analyses assess the generality of our representation findings beyond the 4B configuration and the compact summary in the main paper.

\subsection{t-SNE Across Model Scales}
\label{sec:supp_tsne_scales}

We further extend our representational analysis from the 4B pair to the 8B same-scale setting to verify whether the modality-dependent disparity observed in the main paper holds at larger model capacities. Evaluating matched samples with fixed text continuations across aligned model depths, we find in Figure~\ref{fig:8b_tsne} that VLM and LLM language representations overlap substantially in the Value space while retaining partially separated local structures in the Key space. Conversely, visual representations remain clearly separated from language representations across both Key and Value spaces, even following projection. Consistent with the 4B results, the attention-output visualizations in Figure~\ref{fig:4b8b_attn_tsne} reveal strong overlap among language outputs, whereas visual outputs remain divergent under both native and projected conditions. These observations indicate that the visual-language representational gap is scale-invariant across the tested capacities and extends from cached KV states into subsequent attention outputs.

\begin{figure}[t]
  \centering
  \includegraphics[width=\linewidth]{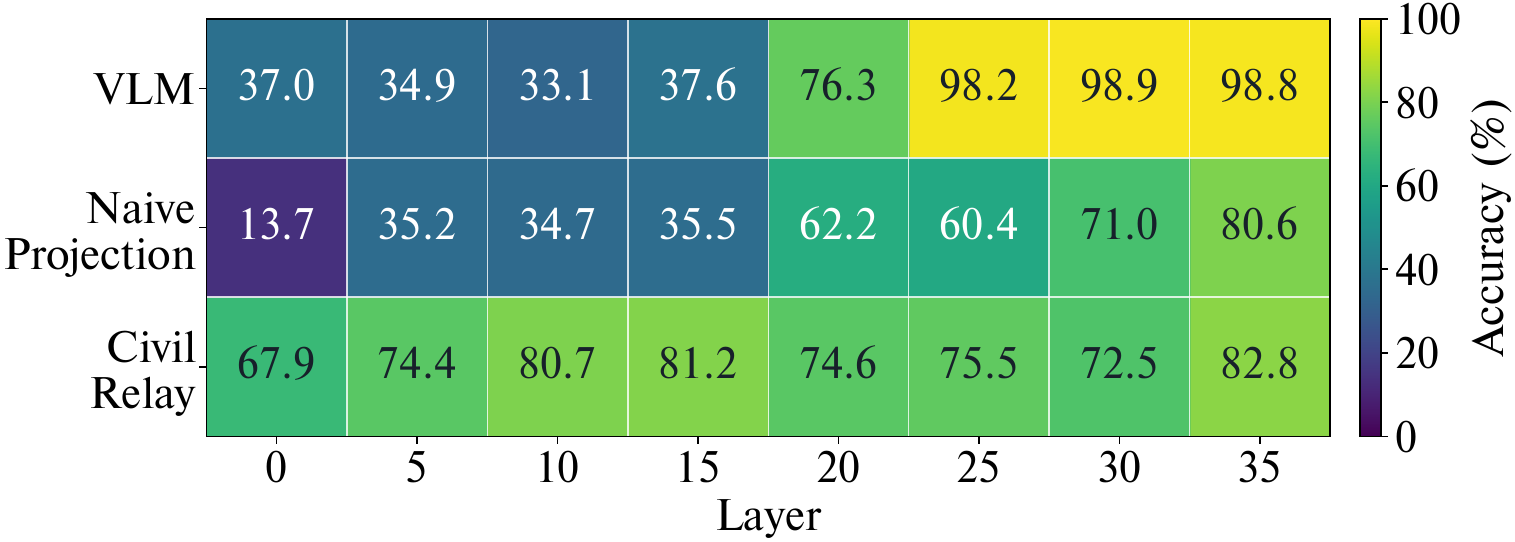}
  \caption{Layer-wise linear probe accuracy on CLEVR. 
  }
  \label{fig:suppl_probe}
\end{figure}
\subsection{CKA across Model Scales}
\label{sec:supp_full_layer_cka}

We extend the CKA analysis with complete groupwise similarity matrices for the 4B$\rightarrow$4B same-scale setting and the 4B$\rightarrow$8B and 8B$\rightarrow$4B cross-scale settings using the same MathVision samples (Figure~\ref{fig:4b8b_cka}). Across all three settings, VLM Decode and Receiver Decode representations exhibit consistently high similarity in both the Key and Value spaces (0.94-0.97). In contrast, both native and projected visual representations show substantially lower similarity to Receiver Decode states, ranging from 0.29 to 0.34 for Keys and from 0.52 to 0.60 for Values. Meanwhile, native and projected visual representations remain highly consistent with each other (0.93-0.96). Taken together, these results reveal a consistent modality-dependent gap across both same-scale and cross-scale communication, indicating that corresponding language representations share closely aligned structures across models whereas visual representations retain their distinct geometry after projection and remain less aligned with Receiver language states. Consequently, cache projection preserves visual representational structure across model scales while leaving a persistent visual-language gap.

\begin{figure*}[t]
    \centering
    \includegraphics[width=\textwidth]{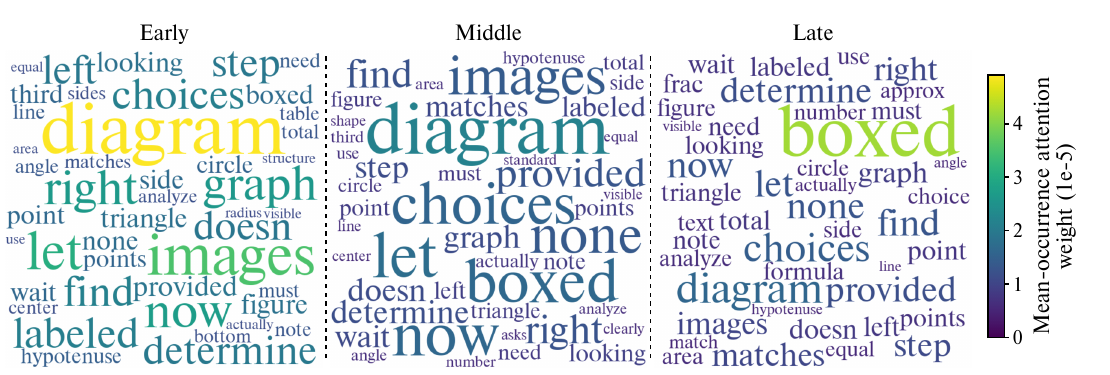}
    \caption{Receiver attention to Planning words during early, middle, and late decoding stages, aggregated across five benchmarks. Each panel displays the 45 words with the highest mean-occurrence attention weights. Early decoding emphasizes visual references and spatial cues, middle decoding combines continued attention to visual content with task-oriented expressions, and late decoding places greater emphasis on answer-related expressions while retaining visual and geometric references. Font size indicates relative weight within each stage and is independently normalized, whereas color reflects a shared attention-weight scale across stages.
    }
    \label{fig:planning_wordcloud}
\end{figure*}

\subsection{Linear Probing on CLEVR}
\begin{figure}[t]
  \centering
  \includegraphics[width=\linewidth]{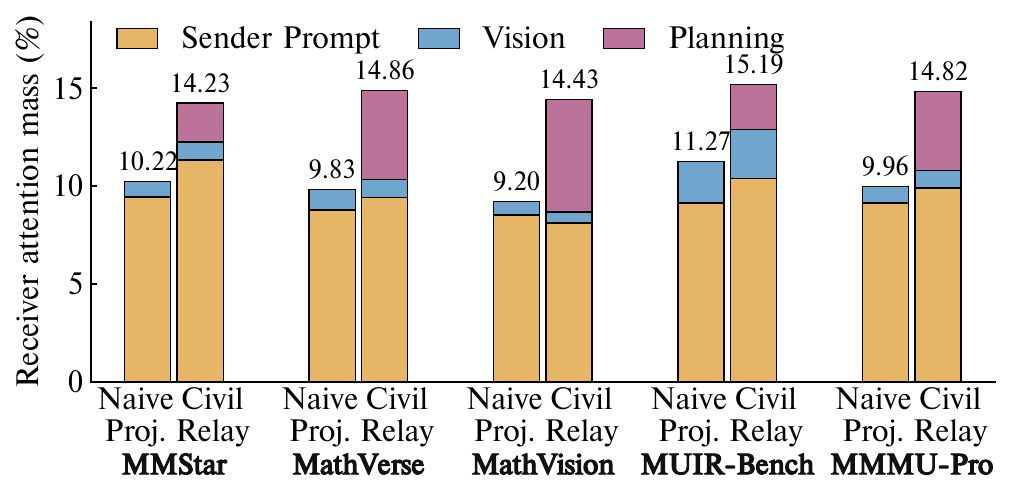}
  \caption{The percentage of the receiver's attention allocated to the sender-related KV cache, averaged over evaluation samples within each benchmark.
  }
  \label{fig:suppl_attnmass}
\end{figure}

To complement the representation-level analysis, we conduct layer-wise linear probing~\cite{alain2016probe} on CLEVR~\cite{johnson2017clevr} to measure how much task-relevant visual information remains linearly decodable. Specifically, for each method and sampled layer, we train a linear classifier to predict the answer color among eight classes from the attention output at the final token of an answer-free prompt. All methods share image-disjoint training, validation, and test sets of 8,000, 1,000, and 2,000 examples, respectively. Only the linear classifiers are trained; the backbone models and projectors remain frozen. Figure~\ref{fig:suppl_probe} compares probing accuracy across layers for the native VLM, Naive Projection, and our Civil-Relay. In deeper layers, native VLM representations achieve high probing accuracy (98.2-98.9\%), whereas Naive Projection reaches 60.4-80.6\%. Across every sampled layer, Civil-Relay outperforms Naive Projection, with the largest gains in shallow and intermediate layers (layers 0-15), where accuracy rises from 13.7-35.5\% to 67.9-81.2\%. In deeper layers, Civil-Relay's advantage narrows as Naive Projection retains more linearly recoverable visual information than at shallow layers. Nevertheless, Civil-Relay remains more accurate at every sampled depth, while both methods fall below the native VLM at these deeper layers. A likely reason is that Planning selectively distills and reorganizes visual evidence relevant to the question rather than preserving every raw visual attribute present in the native VLM, and cross-model projection may further constrain linear decodability. Overall, Civil-Relay provides more consistent access to task-relevant visual information across layers than Naive Projection without attempting to replicate all visual attributes of the native VLM.

\section{What the Receiver Reads During Solving}
\label{sec:supp_attention_content}

While the main paper attention summary quantifies the overall attention allocated to the Sender cache, understanding how cross-model communication supports reasoning requires examining what specific information the Receiver reads during decoding. In this section, we first decompose attention allocation across distinct cache regions, including Sender Prompt, Vision, and Planning, to clarify how direct and indirect visual pathways interact. We then analyze the specific Planning words attended to across early, middle, and late decoding stages to illustrate how the Receiver retrieves visual and task-relevant cues throughout problem solving.

\subsection{Attention Allocation to Transferred KV}
\label{sec:supp_attention_length}

Figure~\ref{fig:suppl_attnmass} decomposes Receiver attention over the transferred cache into Sender Prompt, Vision, and Planning across five benchmarks. Compared with Naive Projection, Civil-Relay raises total Receiver attention to Sender states during decoding from 9.20-11.27\% to 14.23-15.19\%. This additional attention to Planning is not accompanied by a systematic reduction in direct Vision attention, which remains broadly comparable across the two methods. The pattern supports a complementary relationship between the two reading routes. The Receiver continues to access the original Vision KV while additionally attending to Planning states formed under visual context. Planning therefore provides an indirect route to visual evidence alongside direct visual access, rather than simply drawing attention away from the Vision region. Consequently, attention to the Vision region alone does not capture the full reading pattern of Civil-Relay.

Across all benchmarks, Sender Prompt still receives the largest share of attention, exceeding that of Planning and indicating that Planning adds a supplementary reading channel without displacing the original problem context. One plausible explanation is that Sender Prompt acts as a semantic aggregation hub in cache communication, anchoring both visual evidence and generated Planning to the task instructions and serving as a recurring reference as the Receiver integrates information from Vision and Planning. While Planning can reinforce access to this shared context and yield modest increases in Sender Prompt attention on some benchmarks, Prompt attention decreases on others, showing that this shift varies across tasks. Overall, Civil-Relay expands attended Sender memory through Planning while preserving direct visual access and the central contextual role of Sender Prompt.

\subsection{Planning Words Related to Vision}
\label{sec:supp_attention_words}

Figure~\ref{fig:planning_wordcloud} visualizes the Planning words attended to by the Receiver across early, middle, and late decoding stages. The word clouds are constructed via teacher-forced replay of saved Solver responses, with word weights aggregated across samples from five benchmarks. During early decoding, attention prominently targets visual references such as \emph{diagram}, \emph{image}, and \emph{graph}, alongside spatial cues such as \emph{right} and \emph{left}, showing that the Receiver reads Planning expressions describing visual content and spatial relationships. In middle decoding, the Receiver continues to attend to visual references while also reading task-oriented expressions such as \emph{choices}, \emph{find}, and \emph{determine}. In late decoding, attention shifts toward answer-related expressions, most notably \emph{boxed}, while visual and geometric terms persist. This progression illustrates sustained attention to visually related Planning content as decoding moves from problem interpretation toward answer generation, allowing the Receiver to draw on visual evidence without relying solely on the original Vision KV. Rather than acting merely as a preliminary instruction, Planning serves as a persistent source of visual context throughout generation.


\end{document}